\documentclass[conference]{IEEEtran}
\usepackage{times}

\usepackage[numbers]{natbib}
\usepackage{multicol}
\usepackage{hyperref}
\usepackage{url}

\usepackage{graphicx}
\usepackage{diagbox}
\usepackage{multirow}
\usepackage{subfigure}
\usepackage{comment}
\usepackage{placeins}
\usepackage{dblfloatfix}
\usepackage[table]{xcolor}
\usepackage[bottom]{footmisc}

\usepackage[acronym]{glossaries}
\usepackage{amssymb}
\usepackage{tikz}
\usetikzlibrary{bayesnet}
\usepackage{flushend}

\usepackage[ruled,vlined]{algorithm2e}

\usepackage{amsmath,amsfonts,bm}

\def\1{\bm{1}}

\def\vmu{{\bm{\mu}}}
\def\vtheta{{\bm{\theta}}}
\def\va{{\bm{a}}}

\def\vf{{\bm{f}}}
\def\vg{{\bm{g}}}

\def\vq{{\bm{q}}}

\def\vs{{\bm{s}}}

\def\vv{{\bm{v}}}

\def\vx{{\bm{x}}}

\def\vz{{\bm{z}}}

\def\vmu{{\boldsymbol{\mu}}}

\def\vtheta{{\boldsymbol{\theta}}}

\def\vLambda{{\boldsymbol{\Lambda}}}

\def\vSigma{{\boldsymbol{\Sigma}}}

\def\mJ{{\bm{J}}}
\def\mK{{\bm{K}}}

\def\mLambda{{\bm{\Lambda}}}

\DeclareMathAlphabet{\mathsfit}{\encodingdefault}{\sfdefault}{m}{sl}
\SetMathAlphabet{\mathsfit}{bold}{\encodingdefault}{\sfdefault}{bx}{n}

\def\gA{{\mathcal{A}}}

\def\gC{{\mathcal{C}}}
\def\gD{{\mathcal{D}}}

\def\gH{{\mathcal{H}}}

\def\gN{{\mathcal{N}}}

\def\gR{{\mathcal{R}}}
\def\gS{{\mathcal{S}}}

\def\gU{{\mathcal{U}}}

\def\gW{{\mathcal{W}}}
\def\gX{{\mathcal{X}}}

\def\gZ{{\mathcal{Z}}}

\newcommand{\E}{\mathbb{E}}

\newcommand{\R}{\mathbb{R}}

\newcommand{\KL}{D_{\mathrm{KL}}}

\DeclareMathOperator*{\argmax}{arg\,max}
\DeclareMathOperator*{\argmin}{arg\,min}

\newacronym{bc}{BC}{Behavioural Cloning}
\newacronym{il}{IL}{Imitation Learning}
\newacronym{irl}{IRL}{Inverse Reinforcement Learning}
\newacronym{lfd}{LfD}{Learning from Demonstration}
\newacronym{em}{EM}{Expectation Maximization}
\newacronym{promp}{ProMP}{Probabilistic Movement Primitives}
\newacronym{dmp}{DMP}{Dynamic Movement Primitives}
\newacronym{seds}{SEDS}{Stable Estimator of Dynamical Systems}
\newacronym{gmr}{GMR}{Gaussian Mixture Regressor}
\newacronym{gpr}{GPR}{Gaussian Process Regressor}
\newacronym{lwr}{LWR}{Locally Weighted Regressor}
\newacronym{kmp}{KMP}{Kernelized Movement Primitives}
\newacronym{clf}{CLF}{Control Lyapunov Function}
\newacronym{wsaqf}{WSAQF}{Weighted Sum of Asymmetric Quadratic Function)}
\newacronym{nilc}{NILC}{Neurally Imprinted Lyapunov
Candidate}
\newacronym{clfdm}{CLF-DM}{Control Lyapunov Function-based Dynamic Movements}
\newacronym{ciflow}{C-IFlows}{Conditioned ImitationFlows}
\newacronym{iflow}{IFlows}{ImitationFlows}
\newacronym{cnmp}{CNMP}{Conditional Neural Movement Primitives}
\newacronym{tpgmm}{TP-GMM}{Task Parameterized GMM}

\newacronym{gcl}{GCL}{Guided Cost Learning}

\newacronym{mle}{MLE}{Maximun Likelihood Estimation}
\newacronym{sde}{SDE}{Stochastic Differential Equation}
\newacronym{ode}{ODE}{Ordinary Differential Equation}

\newacronym{probs}{ProbS}{Probabilistic Segmentation}
\newacronym{crf}{CRF}{Conditional Random Fields}
\newacronym{ppca}{PPCA}{Probabilistic Principal Component Analysis}
\newacronym{gmcc}{GMCC}{Generalized Multiple Correlation Coeficcient}
\newacronym{hri}{HRI}{Human-Robot Interaction}
\newacronym{ip}{IP}{Interaction Primitives}
\newacronym{hmm}{HMM}{Hidden Markov Model}
\newacronym{cac}{CAC}{Canonical Correlation Coefficient}
\newacronym{rv}{$R_v$}{$R_v$ Coefficient}
\newacronym{dcor}{dCor}{Distance Correlation}
\newacronym{dtw}{DTW}{Dynamic Time Warping}
\newacronym{edr}{EDR}{Edit Distance With Real Penalty}
\newacronym{twed}{TWED}{Time Warp Edit Distance}
\newacronym{r2}{$R^2$}{Coefficient of Determination}
\newacronym{sqp}{SQP}{Successive Quadratic Programming}
\newacronym{rkhs}{RKHS}{Reproducing Kernel Hilbert Space}
\newacronym{icnn}{ICNN}{Input-Convex Neural Network}

\newacronym{pca}{PCA}{Principal Component Analysis}

\newacronym{maf}{MAF}{Masked Autoregressive Flow}
\newacronym{iaf}{IAF}{Inverse Autoregressive Flow}
\newacronym{node}{N-ODE}{Neural ODE}
\newacronym{nsflow}{NSF}{Neural Spline Flows}
\newacronym{cnf}{CNF}{Conditional Normalizing Flows}
\newacronym{ffjord}{FFJORD}{Free-form Jacobian of Reversible Dynamics}

\newacronym{gan}{GAN}{Generative Adversarial Networks}
\newacronym{vae}{VAE}{Variational Autoencoders}
\newacronym{nf}{NF}{Normalizing Flows}
\newacronym{inn}{INN}{Invertible Neural Networks}

\newacronym{rmp}{RMP}{Riemannian Motion Policies}
\newacronym{lmdp}{LMDP}{linearly-solvable Markov Decision Processes}

\newacronym{cep}{CEP}{Composable Energy Policies}

\newacronym{reps}{REPS}{Relative Entropy Policy Search}
\newacronym{rl}{RL}{Reinforcement Learning}
\newacronym{ebm}{EBM}{Energy Based Model}

\newacronym{svi}{SVI}{Structured Variational Inference}
\newacronym{vi}{VI}{Variational Inference}
\newacronym{hrl}{HRL}{Hierarchical Reinforcement Learning}
\newacronym{apf}{APF}{Artificial Potential Fields}
\newacronym{dwa}{DWA}{Dynamic Window Approach}

\begin{document}



\title{Composable Energy Policies for Reactive Motion Generation and Reinforcement Learning}


\author{
    \IEEEauthorblockN{Julen Urain\IEEEauthorrefmark{1}, Anqi Li\IEEEauthorrefmark{2}, Puze Liu\IEEEauthorrefmark{1}, Carlo D'Eramo\IEEEauthorrefmark{1}, Jan Peters\IEEEauthorrefmark{1}\IEEEauthorrefmark{3}}
    \IEEEauthorblockA{\IEEEauthorrefmark{1}IAS, TU Darmstadt,\qquad \IEEEauthorrefmark{2} University of Washington,\qquad \IEEEauthorrefmark{3}MPI for Intelligent Systems},
    \IEEEauthorblockA{Email: \{urain, liu, deramo\}@ias.informatik.tu-darmstadt.de, anqil4@cs.washington.edu,  jan.peters@tu-darmstadt.de}
}


%

\maketitle

\begin{figure}[t]
\begin{minipage}{1\textwidth}
    \includegraphics[width=1\textwidth]{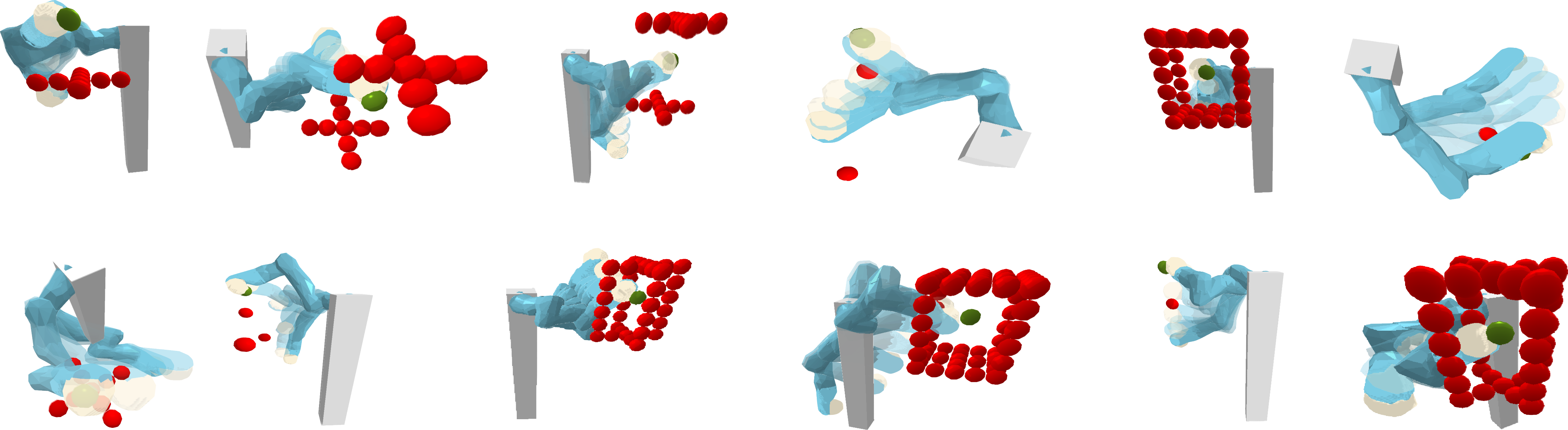}
    \vspace{-0.5cm}
\caption{Composable Energy Policies generate trajectories reactively in a wide set of environments, as the ones in the figure. The generated trajectories are attracted to a target pose, visualized by the green sphere, and avoid a set of obstacles, shown as red spheres.}
\label{fig:teaser}
\end{minipage}
\vspace{-0.5cm}
\end{figure}

\begin{abstract}
Reactive motion generation problems are usually solved by computing actions as a sum of policies. However, these policies are independent of each other and thus, they can have conflicting behaviors when summing their contributions together. We introduce \gls{cep}, a novel framework for modular reactive motion generation. \gls{cep} computes the control action by optimization over the product of a set of stochastic policies. This product of policies will provide a high probability to those actions that satisfy all the components and low probability to the others. Optimizing over the product of the policies avoids the detrimental effect of conflicting behaviors between policies choosing an action that satisfies all the objectives.
Besides, we show that \gls{cep} naturally adapts to the Reinforcement Learning problem allowing us to integrate, in a hierarchical fashion, any distribution as prior, from multimodal distributions to non-smooth distributions and learn a new policy given them. Video in \href{https://sites.google.com/view/composable-energy-policies/home}{https://sites.google.com/view/composable-energy-policies/home}
\end{abstract}

\IEEEpeerreviewmaketitle

\section{Introduction}

Many robotic tasks deal with finding a control action satisfying multiple objectives. An apparently simple task such as watering some plants requires satisfying multiple objectives to perform it properly. The robot should reach the targets (the plants) with the watering can, avoid pouring water on the floor while approaching, and avoid colliding and break plant's branches with its arms. In contrast with more sequential tasks~\cite{sutton1999between, kaelbling2011hierarchical, kaelbling2013integrated, silver2017mastering}, in which the objectives are satisfied concatenating them in time, in the presented work, we consider tasks in which multiple objectives must be satisfied in parallel.

The problem has been faced with a spectrum of solutions that balance between global optimality and computational complexity. Path planning methods~\cite{lavalle2001rapidly, lavalle1999planning, kavraki1996probabilistic} find a global \\
\\
\vspace{0.01cm}

\vspace{5.1cm}
\noindent  trajectory from start to goal by a computationally intense Monte-Carlo sampling process. Trajectory optimization methods~\cite{toussaint2009robot, ratliff2009chomp, kalakrishnan2011stomp, schulman2014motion} reduce the computational burden of planning methods by learning the global trajectory given an initial trajectory candidate. These methods reshape the global trajectory to satisfy the objectives. However, they still require solving an optimization problem over long temporal horizon trajectories. Reactive Motion Generators, such as \gls{apf} methods~\cite{khatib1985potential, koren1991potential, ge2002dynamic, khatib1987unified,ratliff2018riemannian, cheng2018rmpflow} have a very low computational cost, but lack any guarantees of finding a global trajectory satisfying the objectives. These methods propose a modular approach. Each component proposes a deterministic policy to satisfy one of the objectives, and the control action is obtained by the sum of the policy contributions. 

\begin{figure*}[t]
	\centering
	\includegraphics[width=.95\textwidth]{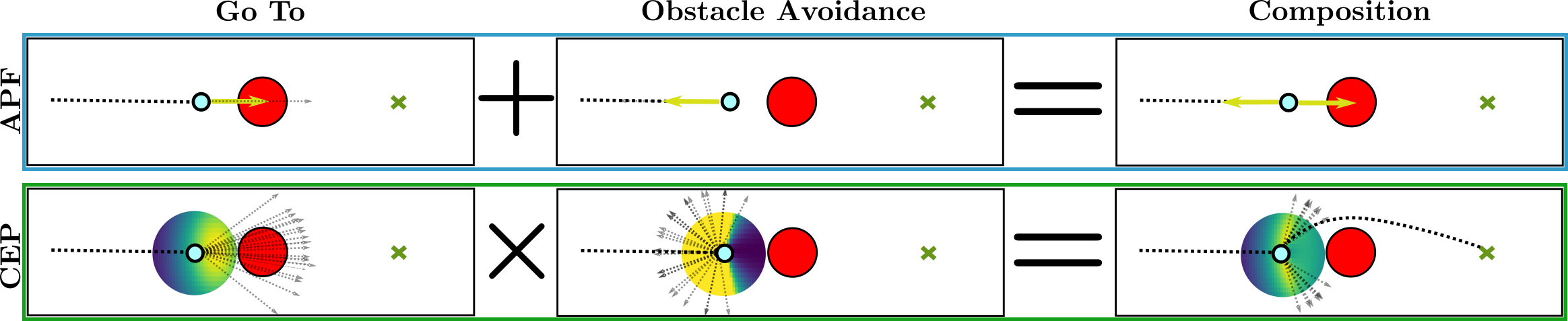}
	\caption{\scriptsize{ Visual Representation of modular control for Goto + Obstacle Avoidance. In the top box, we show Artificial Potential Field~(APF)~\cite{khatib1985potential} policy. In the bottom box, we show Composable Energy Policies (CEP). In contrast, with artificial potential field method, that sums deterministic actions (goto, avoid obstacle), \gls{cep} computes the product of the policy distributions. The composition will provide high probability to those actions that satisfy both components and low to the rest. This approach helps avoiding conflicts between different components. \textit{Robot: blue circle, obstacle: red circle and target: green cross. Thick dotted line: performed trajectory, light dotted line: possible future trajectories.} } }
	\label{fig:comparison}
	\vspace{-0.5cm}
\end{figure*}

There is a clear gap in methodology. The first (path planning and trajectory optimization), approach the problem as an optimization or inference problem over the combined objectives, while the second (reactive motion generation) assumes a complete independence between objectives, lacking any optimality guarantees. We frame Reactive Motion Generation as an optimization problem  over a product of expert policies~\cite{hinton2002training, MCPPeng19, paraschos2013probabilistic, haarnoja2018composable, tasse2020boolean}
\begin{align}
   \va^* =  \argmax_\va  \prod_k \pi_k(\va|\vs).
\end{align}
Most of the previous approaches for reactive motion generation, compute each action maximizing a certain objective independently and then, sum the actions together~\cite{ratliff2018riemannian, khatib1985potential, ge2002dynamic, howard2002mobile}. This independence between policies can end up in conflicting behaviors leading to oscillations or local minima. In our approach, we first compute the product of a set of stochastic policies, and then, we compute the action maximizing the product of them. This approach can be understood as a probabilistic logical conjunction (\textrm{AND} operator)~\cite{du2019compositional, tasse2020boolean} between the stochastic policies (see Fig.~\ref{fig:comparison} for visual representation).

Nevertheless, there are exceptions. \gls{dwa}~\cite{fox1997dynamic,van2011reciprocal} solves the reactive motion generation for a 2D planar robot in a two steps optimization algorithm. First, the search space of possible actions is reduced given a set of constraints. Then, given an objective function, the optimal action is computed solving an optimization problem. In contrast with \gls{dwa}, in our work, we compute the optimal action only through an optimization phase. Additionally, we consider 7-dof robots instead of planar robots and allow to integrate objectives defined in arbitrarily different task spaces. Finally, given the flexibility of our method, we can integrate policies from multiple sources in a single model. We could integrate obstacle avoidance policies with data-driven learned policies, compute a solution by trajectory optimization and add reaction to unexpected situations or integrate priors in a \gls{rl} policy to accelerate learning and ensure safety.

\paragraph*{\textbf{Contribution}}

We present Composable Energy Policies (CEP), a new framework for modular robot control. In contrast with previous methods~\cite{khatib1985potential,ratliff2018riemannian, rimon1992exact} that solves reactive motion generation with a sum of deterministic policies, our method solves reactive motion generation by maximum likelihood estimation over a product of expert policies. We claim that optimizing over the composed policy distributions will increase the guarantees of satisfying all the objectives jointly w.r.t. adding actions coming from different policies. To validate our claims, we derive the reactive motion generation problem from control as inference view and compute the optimality guarantees of applying a product of expert policies (Appendix~\ref{sec:optimality}).

In our work, we aim for flexible composition of modular policy distributions. We propose a framework to integrate stochastic policies represented in arbitrary state-action spaces. Additionally, these policies might have multiple sources from data-driven learned multimodal policies to computationally costly control as inference solutions. In section \ref{sec:transfer}, we show how to integrate a \gls{rl} policy with a set of prior policies to improve exploration and guarantee safety exploration.

In the following, we focus on presenting the method's main blocks while in the Appendix, we focus on the theoretical interpretation of the method. In section \ref{sec:evaluation}, we validate our algorithm in both reactive motion generation and prior based \gls{rl} problems.

\paragraph*{\textbf{Notation}} \textit{As our discussion will involve a set of policies and a set of spaces in which these policies are represented, we will use superscript ($\pi^{x}$) to represent the space in which the policy is and subscript ($\pi_{x}$) to represent the policy index. $\vf_x^z(\cdot)$ represent a transformation map from space $\gX$ to space $\gZ$, $\vs^x$ is the state in space $\gX$ and $\vs^z$, the state in space $\gZ$.}
\section{Preliminaries}

\subsection{Artificial Potential Fields}
Reactive motion generation deals with the problem of generating local and quick motion. The developed methods require to have low computational burden to give a fast response reacting to unexpected situations. \gls{apf}~\cite{khatib1985potential} propose to solve the problem by a weighted sum of a set of dynamic systems represented in a set of task spaces. Let's assume a set of transformations maps $\vf_x^{z_0}, \dots, \vf_x^{z_K}$, that transform a point in the configuration space, $\gX$, to a set of task spaces, $\gZ_k$, $(\vz_k, \dot{\vz_k}) = (\vf_x^{z_k}(\vx), \mJ^{z_k}(\vx)\dot{\vx})$, with $\mJ^{z_k} =\frac{\partial \vf_x^{z_k}(\vx)}{\partial \vx}$; and a set of deterministic second-order dynamic systems in the task space, $\vg^{z_0}, \dots, \vg^{z_k}$. The \gls{apf} represents the dynamic system in the configuration space $\gX$
\begin{align}
    \label{eq:apf_dynamics}
	\vg^{x} = \sum_{k=0}^{K} \mJ^{z_k^{+}}\mLambda_k(\vz_k) \vg^{z_k}
\end{align}
with a weighted ($\mLambda_k(\vz_k)$) sum of the of the projected dynamics and $\mJ^{z_k^{+}}$ the pseudoinverse Jacobian. Each dynamic component represent the policy to satisfy a particular objective. Given that we are looking for the motion that satisfy multiple objectives, the sum of the dynamics might not be enough as the different components might have conflicting behaviours and thus lead to local minima or oscillations. In Appendix~\ref{sec:prove_2}, we show that \gls{apf} can be rewritten as a particular case of product of experts and provide an interpretation of their failures.

We aim to solve the conflicts between dynamics by the maximum likelihood estimation over the product of a set of dynamic distributions. In our work, we model each dynamic component by a distribution. These distributions can be understood as a weighted set of possible dynamics for each component. In the combination of these distributions, our method will search for the dynamics that better satisfy the combined set of objectives and thus, avoid conflicting behaviors, as shown in Figure~\ref{fig:comparison}.

\section{Composable Energy Policies}

Let us assume a set of independent stochastic policies $\pi_1(\va|\vs), \dots, \pi_K(\va|\vs)$ modeled by a Boltzmann distribution
\begin{align}
    \pi_i(\va|\vs) = \exp(E_i(\va,\vs))\frac{1}{Z_i(\vs)}
    \label{eq:ebm_policy}
\end{align}
with $E: \gS \times \gA \rightarrow \R$ is an arbitrarily represented energy function and $Z(\vs) = \int_\va \exp(E(\va;\vs)) d\va$ is the normalization factor. 
Choosing a Boltzmann distribution is not an arbitrary choice. Boltzmann distribution allows representing an arbitrary distribution by a suitable definition of the energy function ($E$)~\cite{lecun2006tutorial}. Additionally, computing the product of experts
\begin{align}
    \pi(\va | \vs) = \prod_{k=0}^{K} \pi_k(\va|\vs)^{\beta_k}\propto \, \exp \left(\sum_{k} \beta_k E_k(\va, \vs)\right).
    \label{eq:product_of_energies}
\end{align}
will end up in a weighted sum over the individual energy components in the log-space. Linearly combined energies are beneficial for computing the energies contribution in parallel and thus, in practise, parallelize the computation by multi-processing, increasing the control frequency. Product of experts is a natural choice for our policy; given a set of energies $E_1,\dots, E_K$, product of experts is the posterior of an inference problem maximizing the energies (See Appendix~\ref{sec:inference_cep}).

\subsection{Energy Tree}
In the composition proposed in \eqref{eq:product_of_energies}, each energy function is considered to be in the same state-action space. However, in most of the robotics scenarios, we might be interested in composing together energies defined in different task spaces. Navigation of the robot's end-effector towards a target while avoiding the obstacles, composes skills defined in different task spaces. Each robot link should avoid obstacles and the end effector should reach a certain target position. Inspired by \gls{apf}~\cite{khatib1985potential} and \gls{rmp}~\cite{ratliff2018riemannian}, we propose to model the composition of energies in different task spaces. We introduce the probabilistic graphical model for our policy in Fig.\ref{fig:energy_tree}.

Our architecture is composed of two main components. First, we have a set of policies $\pi^{z}(\va^z|\vs^z)$, defined in different state-action latent spaces (task space) $(\vs^z, \va^z) \in \gZ$. Second, we consider a set of deterministic mappings that transform the state-actions in the common space (configuration space) $(\vs^x, \va^x) \in \gX$ to the latent state-action spaces $\gZ$, $\vf_{x}^z: \gX \xrightarrow{} \gZ$.

Applying the change of variable rule in probabilistic density functions~\cite{ben1999change, ben2000application}, given the mapping $\vf_{x}^z$ is bijective in the actions ($\va^z \leftrightarrow \va^x$), we can write the policy in the common space, $\pi^{x}$, in terms of the latent space policy $\pi^z$~\cite{urain2020imitation}
\begin{align}
    \pi^x(\va^x|\vs^x) &= \pi^z(\va^z|\vs^z) \sqrt{\det(\mJ^{\intercal} \mJ)}  \nonumber \\
    &= \pi^z(\vf_x^z(\vs^x,\va^x)) \sqrt{\det(\mJ^{\intercal} \mJ)}.
    \label{eq:change_of_variable}
\end{align}
with $\mJ = \partial \vf_{x}^z(\va^x, \vs^x)/\partial \va^x$ is the Jacobian of $\vf_{x}^z$ in the action $\va^x$. The equality in \eqref{eq:change_of_variable} does not hold for injective or surjective transformations. In practice, we apply the equality for any possible map $\vf_{x}^z$ (bijective, surjective, and injective), and we leave for future work the study of the policy distribution under surjective or injective mappings.

From \eqref{eq:change_of_variable}, we can write $\pi^x(\va^x|\vs^x)$ w.r.t. the energy function defined in the latent space $E^z(\va^z ;\vs^z)$
\begin{align}
    \pi^x(\va^x|\vs^x) = \exp \left(E^z(\vf_{x}^z(\va^x,\vs^x))\right)\frac{1}{Z(\vs^x)}
    \label{eq:policy_change}
\end{align}
with 
\begin{align}
    Z = \frac{\int_{\va^z} \exp(E^z(\va^z,\vs^z)) d\va^z}{\sqrt{\det(\mJ^{\intercal}\mJ)}}.
\end{align}
This policy representation allows to define the energy function in an easy to represent task space. Then, for any $\vs^x, \va^x \in \gX$, we can map them to the latent space $\gZ$, $(\vs^z , \va^z) = f_{x}^{z}(\vs^x, \va^x)$ and compute the un-normalized log-probability of the action in a given latent policy $E^z(\va^z,\vs^z)$. 

Our energy tree is inspired by the \gls{apf} formulation, but it has important differences. In \gls{apf} formulation, there is an explicit function from state $\vs^x$ to action $\va^x$
\begin{align}
    \va^x = \vf_{x}^{z-1}(\va^z) = \vf_{x}^{z-1}(\vg(\vs^z)) = \vf_{x}^{z-1}(\vg(\vf_{x}^z(\vs^x)))
    \label{eq:explicit_dyn}
\end{align}
where $\vf_{x}^z$ is the function mapping from common space to latent space and $\vg$ is a deterministic policy in the latent space. In this approach, $\vf_{x}^z$ is required to be invertible. In contrast, \gls{cep} assumes an implicit function from $\vs^x$ to $\va^x$
\begin{align}
    \va^{x} = \argmax_{\va^x} E^z(\vf_x^z(\va^x, \vs^x)).
\end{align}
In this approach, we do not need to compute the inverse of $\vf_x^z$. Additionally, $E^{z}$ represents the energy in the latent space, so we can combine distributions instead of combining deterministic dynamics. In Sec.~\ref{sec:robot_motion} we show that having an implicit function might be beneficial for reactive motion generation.

In case a set of policies exist in different task spaces $\gZ_1, \dots, \gZ_N$, we can compose the policies together 
\begin{align}
	\label{eq:cep_generic}
    \pi^x(\va^x|\vs^x) = \exp \left(\sum_{k=1}^{K}\beta_k E^{z_k}(\vf_{x}^{z_k}(\va^x,\vs^x))\right) \frac{1}{Z(\vs^x)}
\end{align}
Shown in Fig.~\ref{fig:energy_tree}, \gls{cep} has a recursive architecture. Our model allows us to represent the latent policies $\pi^z(\va^z|\vs^z)$ with a set of policies in an even deeper latent space $\gW$, and still compute the optimization in the same way.

\begin{figure}[t]
	\centering
	\resizebox{0.5\textwidth}{!}{\input{img/models/energy_tree}}
	  	
  \caption{Bayesian network for the energy tree. The policy distribution $\pi^x(\va^x|\vs^x)$ is computed w.r.t. the policies in the latent spaces $\gZ_i$. Additionally, a latent policy $\pi^z(\va^z|\vs^z)$ can be represented by a set of policies in a different latent spaces $\gW_i$. \textit{White diamonds represent deterministic transformations and black square is a probabilistic mapping.}}
  \label{fig:energy_tree}
\end{figure}

\subsection{Maximum Likelihood Estimation on CEP}
\label{sec:optimization_algorithm}
In \gls{cep}, the optimal action is obtained by Maximum Likelihood Estimation over the product of expert policies. Given the current state $\vs^x$
\begin{align}
	\label{eq:optim_cep}
	\va_{\textrm{ML}}^x = \argmax_{\va^x} \prod_{k=0}^{K}\pi_k(\va^x|\vs^x)^{\beta_k} \, ,\, \beta_k>0.
\end{align}
We  assume a set of mappings $\vf_{x}^z$ and a set of energies $E^{z}$ are given. We solve \eqref{eq:optim_cep}, by Cross Entropy optimization~\cite{rubinstein1999crossentropy}. We define a proposed sampling model $p(\va|\vtheta)$ and iteratively solve
\begin{align}
	\vtheta_{t+1} &= \argmax_{\vtheta_t} \E_{p(\va^x|\vtheta_t)} \left[ \log\left(\prod_{k=0}^{K} \pi_k(\va^x|\vs^x)^{\beta_k}\right)\right] \nonumber \\
	 & \propto \E_{p(\va^x|\vtheta_t)} \left[ \sum_{k=1}^{K}\beta_k E^{z_k}(\vf_{x}^{z_k}(\va^x,\vs^x)) \right].
\end{align}
The method is presented in Algorithm~\ref{alg:cep}. Each energy ($E^z$) component is linearly dependent and thus, in the optimization process, we evaluate each energy component in parallel. Also, we consider a batch $N$ of actions evaluated all in parallel on GPU. This parallelization reduces the computational time for the optimization process and allows to use \gls{cep} for high-frequency reactive motion generation.

\begin{algorithm}
	\SetKwInOut{Input}{Given}
	\Input{$N$: Number of samples\;
		$\vs^x$: Current state in common space\;
		$K$: Number of Energy Components\;
		$(\vf_x^{z_1}, E^{z_1},\beta^{z_1}), \dots, (\vf_x^{z_K}, E^{z_K}, \beta^{z_K}))$: Maps, energies and inverse temperatures\;
		$I$: Optimization steps\;
		$(\mu_0, \Sigma_0)$: Initial sampling distribution mean and variance\;
		$(\va^{*}, e^{*})$: Initial optimal action and energy\;}
	\BlankLine
	\For{$i \leftarrow 0$ \KwTo $I-1$}{
		
		\For{$n \leftarrow 0$ \KwTo $N-1$}{
			$\va_n^{x} \sim \gN(\mu_i, \Sigma_i)$\;
			\For{$k \leftarrow 1$ \KwTo $K$}{
				$\vs^{z_k},\va_n^{z_k} = \vf_x^{z_k}(\vs^x, \va_n^{x})$\;
				$e_n^{z_k} = E^{z_k}(\vs^{z_k},\va_n^{z_k})$\;
			}
			$e_n^x = \sum_{k=1}^{K}\beta^{z_k}e_n^{z_k}$
		}
		\BlankLine
		$\mu_{i+1} \leftarrow \textrm{Update}_\mu(\mu_{i},\va_{0:N}^{x} ,e_{0:N}^x)$\;
		$\Sigma_{i+1} \leftarrow \textrm{Update}_\Sigma(\Sigma_{i},\va_{0:N}^{x} ,e_{0:N}^x)$\;
		\BlankLine
		$\va_i^{*}, e_i^* \leftarrow \arg_\va \max_e (e_{0:N}^x)$\;
		\If {$e^* <e_i^*$}{
			$\va^* \leftarrow \va_i^{*}$\;
			$e^* \leftarrow e_i^*$\;
		}

	}
	return $\va^*$\;
	
	\caption{Composable Energy Policies \label{alg:cep}}
\end{algorithm}

\section{CEP for Robot Motion Generation}
\label{sec:robot_motion}

In this section, we will provide further insights on how to model the mapping functions and the stochastic policies for a robot motion generation problem. We consider the motion is generated with a second-order dynamic system in the robot's configuration space
\begin{align}
	\label{eq:sec_ord_fun}
	\ddot{\vq} = g(\vq,\dot{\vq})
\end{align}
where $\vq$ is the robot's joint state, $\dot{\vq} = d\vq / d t$ velocity and $\ddot{\vq} = d^2\vq / dt^2$, the acceleration.

Our dynamic system is modeled by the maximization in \eqref{eq:optim_cep}. Thus, we represent our action $\va^x = \ddot{\vq}$ and the state $\vs^x=\{\vq, \dot{\vq}\}$.

The energy distributions are defined in a set of task spaces. We model the map between the common space (configuration space) and the latent space (task space) $\vf_{q}^{x}$ by the robot's kinematics
\begin{align}
	\vx &= \textbf{f}_{\textrm{kin}}(\vq) \nonumber \\
	 \label{eq:kinematics}
	\dot{\vx} &= \mJ(\vq)\dot{\vq}\\
	\ddot{\vx} &= \mJ(\vq)\ddot{\vq} + \dot{\mJ}(\vq)\dot{\vq} \approx  \mJ(\vq)\ddot{\vq} \nonumber
\end{align}
with forward kinematics $\textbf{f}_{\textrm{kin}}$ to a given task space and $\mJ(\vq) = \partial \vx / \partial \vq$, the Jacobian for the given forward kinematics.

For the particular case in which the mappings are given by \eqref{eq:kinematics}, we represent the general policy in \eqref{eq:cep_generic} as
\begin{align}
	\label{eq:reactive_cep}
    \pi^q(\ddot{\vq}|\vq, \dot{\vq}) \propto \exp \left(\sum_{k=1}^{K}\beta_k E^{x_k}(\mJ_k(\vq)\ddot{\vq},\textbf{f}_{\textrm{kin}k}(\vq), \mJ_k(\vq)\dot{\vq}) \right).
\end{align}
We remark, that computing the optimal solution for \eqref{eq:reactive_cep}, given we use an implicit representation, we only require access to the forward kinematics $\textbf{f}_{\textrm{kin}}$ and the Jacobian $\mJ$ functions. Instead, explicit methods~\cite{ratliff2018riemannian, khatib1985potential}, require to compute the Jacobian pseudoinverse to get the action in the configuration space as shown before in \eqref{eq:apf_dynamics} and \eqref{eq:explicit_dyn}. Jacobian pseudoinverse might be problematic to compute if the robot is in singularities.

Energy functions allow having a flexible policy distribution representation. In the following, we provide some energy proposals for the main components the robot requires for reactive motion generation. Additionally, we show in Appendix~\ref{sec:prove_2}, that the weighted sum of deterministic policies (Artificial Potential Fields methods), can be framed as a particular case of \gls{cep}.

\paragraph*{\textbf{Go To a target}}
The simplest distribution model for going to a target position is a normal distribution
\begin{align}
	\pi(\ddot{\vx}|\vx, \dot{\vx}) = \gN(\vmu(\vx,\dot{\vx}), \vSigma(\vx))
\end{align}
with
\begin{align}
	\vmu(\vx,\dot{\vx}) = - \mK_p(\vx - \vx^*) - \mK_v \dot{\vx} \\
	\vSigma(\vx) = \alpha(\vx - \vx^*)^{\intercal}(\vx - \vx^*).
\end{align}
The distribution proposes as mean $\vmu$, a PD-controller with $\vx^*$ the target state and $\mK_p>0$ and $\mK_v>0$ gains and a covariance matrix $\vSigma$ represented as the quadratic distance to the target, scaled by $\alpha>0$. The variance will shrink closer to the target and, thus, the relevance of this energy component will increase the closer we are to the desired state.

\paragraph*{\textbf{Obstacle Avoidance}}
We represent obstacle avoidance energy in the unidimensional space represented by the vector between the cartesian robot position $\vx_r$ in task space and the cartesian obstacle position $\vx_o$.

We first compute the vector pointing to the obstacle
\begin{align}
	\vv_{r,o} = \vx_r - \vx_o
\end{align}
Then, we project the velocity $\dot{\vx}$ and the acceleration $\ddot{\vx}$ vectors in the task space
\begin{align}
	\dot{x}_p &= \dot{\vx} \frac{\dot{\vx} \cdot 	\vv_{r,o} }{||\dot{\vx}|| ||\vv_{r,o}||} \nonumber \\
	\ddot{x}_p &= \ddot{\vx} \frac{\ddot{\vx} \cdot 	\vv_{r,o} }{||\ddot{\vx}|| ||\vv_{r,o}||}
\end{align}
Finally, we compute the energy for the acceleration $\ddot{x}_p$, $	\dot{x}_p$ and $\vv_{r,o}$ as
\begin{align}
	E(\ddot{x}_p, \dot{x}_p, \vv_{r,o}) = \begin{cases}
		0 & \text{\textrm{if} $\dot{x}_p < 0$ \textrm{or} $||\vv_{r,o}|| > \gamma$ } \\
		-\infty & \text{\textrm{if} $\ddot{x}_p > -\alpha\dot{x}_p -\beta$}\\
		0 & \textrm{otherwise}
	\end{cases}       
\end{align}
with a distance threshold parameter $\gamma>0$ to activate the energy component. If the projected robot velocity is not pointing to the obstacle or if the robot's task space position is too far from the robot, the distribution is a uniform distribution over the whole acceleration space. In case the robot is close to the obstacle and the velocity vector points to the obstacle, the distribution then becomes a uniform distribution between $-\infty$ and $-\alpha\dot{x}_p -\beta$, $\gU_p(-\infty,-\alpha\dot{x}_p -\beta)$ with $\alpha >0$ and $\beta>0$.

\paragraph*{\textbf{Avoid Joint Limits}}
Similarly to the collision avoidance energy, we model the joint limits avoidance with a uniform distribution that provides high probability to those accelerations pointing in the opposite direction to the joint limit as long as the joint is close to the limits. Given the distance to the joint limit
\begin{align}
	d_l = q - q_l
\end{align}
with $q_l$ the joint limit, the energy can be represented as
\begin{align}
	E(\ddot{q}; \dot{q}, q) = \begin{cases}
		0 & \text{\textrm{if} $\frac{d_l}{|d_l|}\dot{q} < 0$ \textrm{or} $|d_l| > \gamma$ } \\
		-\infty & \text{\textrm{if} $\frac{d_l}{|d_l|}\ddot{q} > -\alpha\frac{d_l}{|d_l|}\dot{q} -\beta$}\\
		0 & \text{otherwise}
	\end{cases}
\end{align}
with $\alpha>0$ and $\beta>0$.

\paragraph*{\textbf{Imitation Learning, Control-as-Inference and RL}}

\gls{cep} architecture is not only limited to heuristic energies. The architecture allows to integrate any form distributions. \gls{cep} architecture allows to integrate multimodal policies learned by behavioural cloning~\cite{urain2020imitation, paraschos2013probabilistic}, posterior distributions computed by control as inference~\cite{lambert2020stein, watson2020stochastic, williams2017model} or a \gls{rl} policy.

In the following, we propose a novel hierarchical reinforcement learning policy to integrate an \gls{rl} agent with a set of prior distributions.

\section{CEP for Hierarchical Reinforcement Learning}
\label{sec:transfer}
Priors have been widely used in RL, accelerating the learning process and therefore improving the sample efficiency. Most popular approaches can be categorized in two groups. On the one hand, methods that first approximate a parameterized policy to mimic the priors
\begin{align*}
	\min_{\theta} \KL(\pi_{\theta}(\va|\vs) || \pi_{q}(\va|\vs))
\end{align*}
 and then, update this policy to maximize for a new task~\cite{mulling2013learning, urain2020imitation}. 
\begin{align*}
     \max_{\theta} \E_{p_{\pi_{\theta}}(\vs,\va)}[\gR(\vs,\va)].
\end{align*}
Even if this approach is the most popular one, if the parameterized policy is not sufficiently expressive is not going to cover all the prior information. Moreover, the parameterized model might forget important information coming from the prior after some \gls{rl} updates. On the other hand, \gls{hrl} methods~\cite{johannink2019residual, silver2018residual, singh2020parrot, daniel2012hierarchical, MCPPeng19} consider a two-layered policy.
\begin{align}
    \pi^L(\va|\vs) = \int_{\va_H} \pi_q(\va|\vs, \va_H)\pi_{\theta}^H(\va_H|\vs) d \va_H
\end{align}
Given some priors, a policy $\pi_q$ is modeled conditioned on a higher-order action $\va_H$. For sampling an action, we first sample an action from the high-level policy, $\va_H\sim\pi_{\theta}^H(\va_H|\vs)$ and then, we sample from the conditioned low-level policy, $\va\sim\pi_q(\va|\vs, \va_H)$. Most of the previous \gls{hrl} policies assume rather a mixture of skills~\cite{MCPPeng19, heess2016learning, mulling2013learning, florensa2017stochastic} or a simple sum of the prior actions~\cite{johannink2019residual, silver2018residual} as \gls{hrl} policies. These policy architectures might not satisfy the desired properties in a real robot \gls{rl} problem, rather because they are not safe enough or because they are so limited in their exploration space.

In a robot learning scenario, a desired \gls{hrl} policy should consider (1) priors that guarantee hard constraints to avoid collisions, (2) multimodal/non-normal priors to provide a rich set of possible exploring regions, (3) allow to integrate multiple prior sources in the same policy and (4) do not limit the exploration region to a few set of given skills. In the following, we show that \gls{cep} allows integrating all the desired properties in the \gls{hrl} policy.

In our work, we propose to model the low level policy with a \gls{cep}
\begin{align}
	\label{eq:hrl_policy}
	\vmu_H, \vSigma_H &\sim \pi_{\theta}^H(\vmu_H, \vSigma_H|\vs) \\
    \pi^L(\va|\vs, \vmu_H, \vSigma_H) &= \pi_q(\va|\vs)\gN(\va|\vmu_{H},\vSigma_{H})\nonumber
\end{align}
In our model, the action is computed by the maximization over the low-level policy
\begin{align}
	\va = \argmax_{\va} \pi^L(\va|\vs, \vmu_{H}, \vSigma_H).
\end{align}
Our proposed policy combines the given prior policy distribution with a normal distribution parameterized with the high-level policy. This structure allows integrating any type of prior distribution with a learnable policy. The prior could impose hard constraints to avoid actions that lead to collisions by setting $\pi_q(\va|\vs)=0$ or encourage the robot to explore in a mixture of regions by setting a mixture of distributions as prior.

The variance $\vSigma_{H}$ control  controls the importance of the learning policy w.r.t. the priors. The bigger $\vSigma_{H}$ is the less relevant the learning policy is and thus, the higher the influence of the priors in the action; in the limit  $\lim_{\vSigma_H \rightarrow 0} \pi^L(\va|\vs, \vmu_{H}, \vSigma_H) = \delta(\vmu_{H} - \va)$, the prior effect disappear and we only trust in the learning policy.

It is interesting to remark that our model applies \gls{rl} in the high-level policy and solves a stochastic search optimization problem in the low-level policy. With the low-level stochastic search optimization, we can guarantee that some conditions (obstacle avoidance) will be satisfied in the robot action, while, a simple residual control could not.

\section{Experimental Evaluation}
\label{sec:evaluation}
The model evaluation is split into two parts. First, we study \gls{cep} performance for reactive motion generation. We evaluate how the different energy components perform by observing the success rate and the collisions in a set of obstacle avoidance environments. We compare \gls{cep} w.r.t. previous modular reactive controllers. We exclude trajectory optimization and path planning algorithms as our evaluation is interested in reactive controllers that can provide a control response in high control frequencies ($\geq$500Hz).

Second, we evaluate \gls{cep} in an \gls{rl} problem. We want to observe if using \gls{cep} as prior boosts the learning performance of an \gls{rl} agent. Additionally, we want to evaluate if the proposed priors are sufficiently strong to impose safety priors in the policy and avoid collisions. We compare our approach to previous methods applying both \textit{Behavioural Cloning + RL} and \textit{Hierarchical RL}.

\begin{figure}[t]
	\centering
	\includegraphics[width=.4\textwidth]{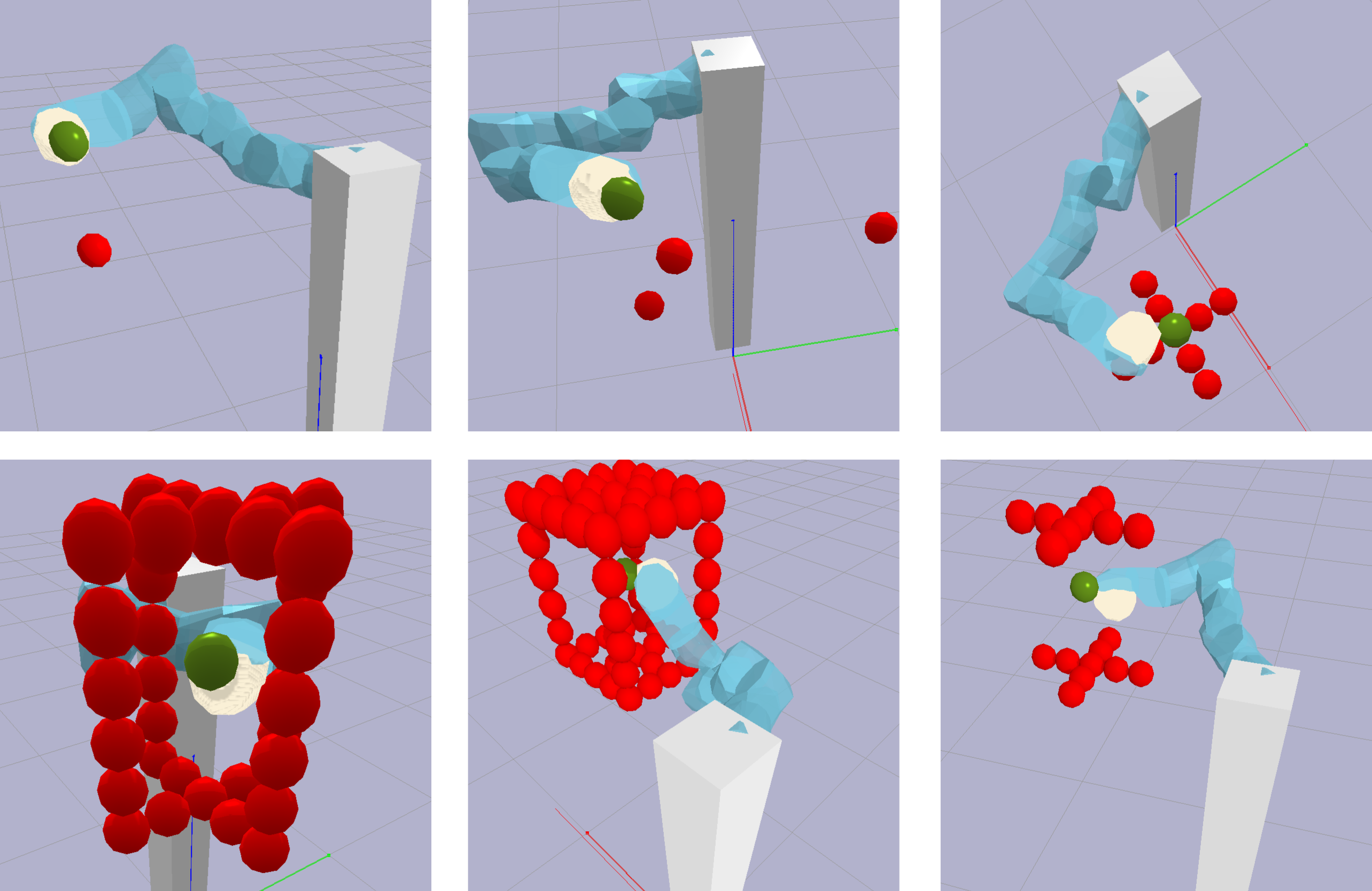}
	\caption{\scriptsize{Reactive Motion Generation Environments. In the top, from left to right: 1 Obstacle, 3 Obstacles, Cross. In the bottom, from left to right: Cage I, Cage II, Double Cross. \textit{Green Sphere: target pose, red spheres: obstacles}.}}
	\label{fig:Environments}
\end{figure}

\subsection*{\textbf{Reactive Motion Generation}}
We consider the problem of controlling a robot manipulator in a set of obstacle avoidance environments. We use a 7 dof KUKA-LWR robot manipulator and the objective is to reach a target pose (3D and 6D) while avoiding colliding with the obstacles. We consider increasingly difficult environments presented in Fig.~\ref{fig:Environments}. The robot is initialized in a random joint configuration and its motion is generated by a set of modular reactive motion generators, without additional global path planning algorithms. We evaluate the success rate and the collision rate of our method and compare it w.r.t. deterministic modular reactive motion controllers (Artificial Potential Fields~\cite{khatib1985potential} and Riemannian Motion Policies~\cite{ratliff2018riemannian}) as baselines.

We build the \gls{cep} model with
\begin{itemize}
	\item Go-To energy in the end effector's task space.
	\item Obstacle Avoidance energy in each robot link's task space.
	\item Joint limits avoidance energy in the configuration space.
\end{itemize}
To guarantee high frequency control actions ($\geq 500Hz$), we solve the optimization problem \eqref{eq:optim_cep} in a Nvidia GPU RTX-2800. Each energy component is parallelized and the action samples are evaluated in batch. With a sufficient amount of samples per optimization step (10.000 samples), the optimization is solved in two steps.

\begin{table*}[t]
	\resizebox{\textwidth}{!}{
	\centering
	\begin{tabular}{ c c c c c c c c c c c c c} 
		\hline
		Methods & \multicolumn{2}{c}{1 Obstacle} &  \multicolumn{2}{c}{3 Obstacles} &  \multicolumn{2}{c}{Cross} &  \multicolumn{2}{c}{Double Cross}& \multicolumn{2}{c}{Cage I} & \multicolumn{2}{c}{Cage II}\\
		& Success & Collide & Success & Collide & Success & Collide & Success & Collide & Success & Collide & Success & Collide\\
		\hline
		Riemannian Motion Policies~\cite{ratliff2018riemannian} & 100/100 & 0/100 & 98/100 & 0/100 & 94/100 & 0/100 &84/100 & 0/100 &55/100&0/100&5/100&0/100\\
		Artificial Potential Fields~\cite{khatib1985potential} & 100/100 & 0/100 & 100/100 & 0/100 & 90/100 & 0/100 & 76/100 & 0/100 &43/100 & 0/100 & 4/100 & 2/100\\
		\textbf{Composable Energy Policies}(Ours) & \textbf{100/100} & \textbf{0/100} & \textbf{100/100} & \textbf{0/100} & \textbf{98/100} & \textbf{0/100} &\textbf{91/100} & \textbf{0/100} &\textbf{89/100} & \textbf{0/100} &\textbf{46/100} & \textbf{0/100} \\
		\hline \\
		
	\end{tabular}
	}
	\caption{Results for 3D GoTo + Obstacle Avoidance Task}
	\label{tab:3d}
\end{table*}

\begin{table*}[t]
	\centering
	\resizebox{\textwidth}{!}{
		\begin{tabular}{ c c c c c c c c c c c c c} 
			\hline
			Methods & \multicolumn{2}{c}{1 Obstacle} &  \multicolumn{2}{c}{3 Obstacles} &  \multicolumn{2}{c}{Cross} &  \multicolumn{2}{c}{Double Cross}& \multicolumn{2}{c}{Cage I} & \multicolumn{2}{c}{Cage II}\\
			& Success & Collide & Success & Collide & Success & Collide & Success & Collide & Success & Collide & Success & Collide\\
			\hline
			Riemannian Motion Policies~\cite{ratliff2018riemannian} & 100/100 & 0/100 & 91/100 & 0/100 & 75/100 & 0/100 & 63/100 &0/100 & 18/100&0/100&1/100&0/100\\
			Artificial Potential Fields~\cite{khatib1985potential}  & 100/100 & 0/100 & 90/100 & 0/100 & 78/100 & 0/100 & 60/100 &0/100 & 21/100 & 0/100 & 1/100 & 0/100\\
			\textbf{Composable Energy Policies}(Ours) & \textbf{100/100} & \textbf{0/100} & \textbf{100/100} & \textbf{0/100} & \textbf{91/100} & \textbf{0/100} &\textbf{87/100} & \textbf{0/100} &\textbf{43/100} & \textbf{0/100} &\textbf{14/100} & \textbf{0/100} \\
			\hline \\
		\end{tabular}}
	\caption{Results for 6D GoTo + Obstacle Avoidance Task}
	\label{tab:6d}
\end{table*}

\paragraph*{\textbf{Results and analysis}}
We summarize the obtained results for the 3D Goto problem in table~\ref{tab:3d}. Few obstacles environments are easily solved by all the methods. We observe a success rate of almost 100\% for the first three environments in the three cases. This result shows that simple scenarios can be easily solved with local reactive controllers and it is not required to solve a global trajectory planning problem. In complex scenarios, \gls{cep} performs better than the baselines. We can obtain an 89\% success rate in the first cage environment and 46\% in the second cage. We think that these results might be related to the way we compute the robot's acceleration. While the baselines methods compute the accelerations provided by the different components(Go-To, Avoid Joint Limits, Obstacle Avoidance), and do not take into account any possible conflict of interests between each action, our method first computes the distribution obtained by the product of all the components and then, computes the action that maximizes this composed distribution. The built composition will provide high probabilities to those actions that have high probability in each component and low to the rest. Thus, the optimal action from the composed distribution is expected to be the one that better satisfies all the individual components.

The performance worsens for the 6D GoTo problem (Table~\ref{tab:6d}). The orientation sets an important constraint in the possible final configurations and reduces the set of trajectories that solves the problem. \gls{cep} was able to perform relatively better than the baselines but it got less than 50\% success rate in both cage environments, suggesting that in complex scenarios an additional global path planner should be integrated with \gls{cep}.

\begin{figure}
	\centering
	\includegraphics[width=.45\textwidth]{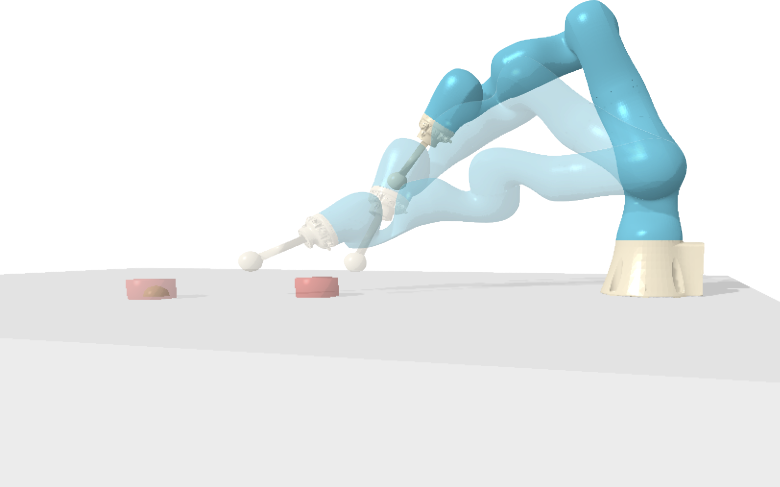}
	\caption{Puck hitting environment. \gls{cep} is tested in a \gls{rl} environment where the robot needs to learn how to hit the puck. The reward is defined by the euclidean distance between the puck and the green target. An additional reward penalizes if the robot hits the table.  }
	\label{fig:hit_env}
	\vspace{-0.5cm}
\end{figure}
\subsection*{\textbf{Prior Based Policies for RL}}

In the following experiment, we evaluate the performance of \gls{cep} as a policy that combines prior policies with a new learnable policy to solve a new task. We consider the problem of hitting a puck and place it in a target position Fig.~\ref{fig:hit_env}. We use a 7 dof LBR-IIWA robot, the action space is defined by the end-effector's task space cartesian velocity $\va \in \R^3$ and the state $\vs \in \R^{18}$ is represented by the end-effector's cartesian position $\vx_{\textrm{ee}}$, puck's position $\vx_{\textrm{puck}}$, their relative position $r_{\textrm{p-ee}} = \vx_{\textrm{ee}} - \vx_{\textrm{puck}}$, the puck's velocity $\vv_{\textrm{puck}}$, the end effector's velocity $\vv_{\textrm{ee}}$ and the target position $\vx_{\textrm{target}}$. One episode has 1400 steps. We define the reward as
\begin{align}
	r = -|| \vx_{\textrm{puck}} - \vx_{\textrm{target}}||^2 + r_T(\vx_{\textrm{ee}})
\end{align}
with $r_T(\vx_{\textrm{ee}}) = -1000$ if the robot's end effector position $\vx_{\textrm{ee}}$ collides the table. The first term $-|| \vx_{\textrm{puck}} - \vx_{\textrm{target}}||^2$ will encourage the robot to put the puck close to the target, minimizing the euclidean distance. The second term will encourage the robot to avoid to collide against the table. We compare \gls{cep} approach presented in \eqref{eq:hrl_policy} with other approaches that combines prior information with \gls{rl}. We consider Behavioural Cloning + \gls{rl}~\cite{mulling2013learning} and Residual Policy Learning~\cite{johannink2019residual, silver2018residual} as baselines.

To guarantee a fair comparison, we apply the same prior policy in the three scenarios. We build a prior policy with two components in a \gls{cep}. We consider
\begin{itemize}
	\item Go-To energy in the robot's end effector's task space.
	\item Table avoidance in the robot's end effector's task space.
\end{itemize} 
The prior is combined in different forms depending on the applied algorithm:

\textit{Behavioural Cloning + RL}. We first fit a policy to the prior policy
\begin{align}
	\vtheta^* = \argmin_{\vtheta} \E_{p(\vs)}\left[\KL(\pi_{q}(\va|\vs)||\pi_{\theta}(\va|\vs))\right]
\end{align}
and then directly sample from the fit policy
\begin{align}
	\va \sim \pi_{\theta}(\va|\vs).
\end{align}

\textit{Residual Policy Learning}. We consider a hierarchical structure. We first sample, from the learnable policy and add the optimal solution from the prior
\begin{align}
	\va = \va^{*} + \va_H \, , \, \va_H \sim \pi_{\theta}(\va_H|\vs)
\end{align}
with
\begin{align}
	\va^{*} = \argmax_{\va} \pi_{\vq}(\va|\vs).
\end{align}

\textit{Composable Energy Policies}. We also consider a hierarchical structure. In this case, we first sample from the learnable policy 
\begin{align}
	\va_H \sim \pi_\vtheta(\va_H|\vs)
\end{align}
and then, optimize over the composition
\begin{align}
	\va = \argmax_{\va}\pi_{q}(\va|\vs,\va_H).
\end{align}
We apply the hierarchical policy introduced in \eqref{eq:hrl_policy}. For all the experiments, we used the Mushroom-RL~\cite{deramo2020mushroomrl} PPO~\cite{schulman2017proximal} implementation. We include additional details in Appendix~\ref{sec:experiments}.

\begin{figure}
	\centering
	\begin{minipage}{.24\textwidth}
		\centering
		\includegraphics[width=.99\textwidth]{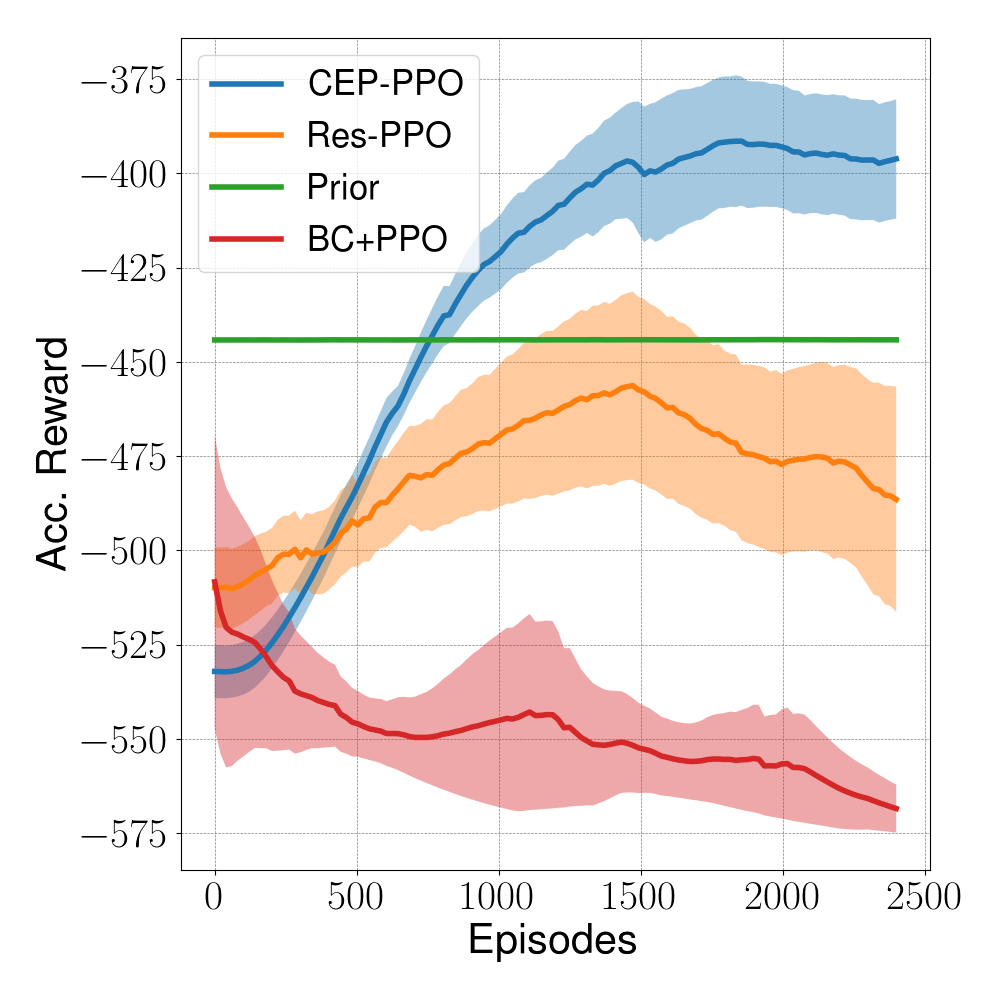}
	\end{minipage}
	\begin{minipage}{.24\textwidth}
		\centering
		\includegraphics[width=.99\textwidth]{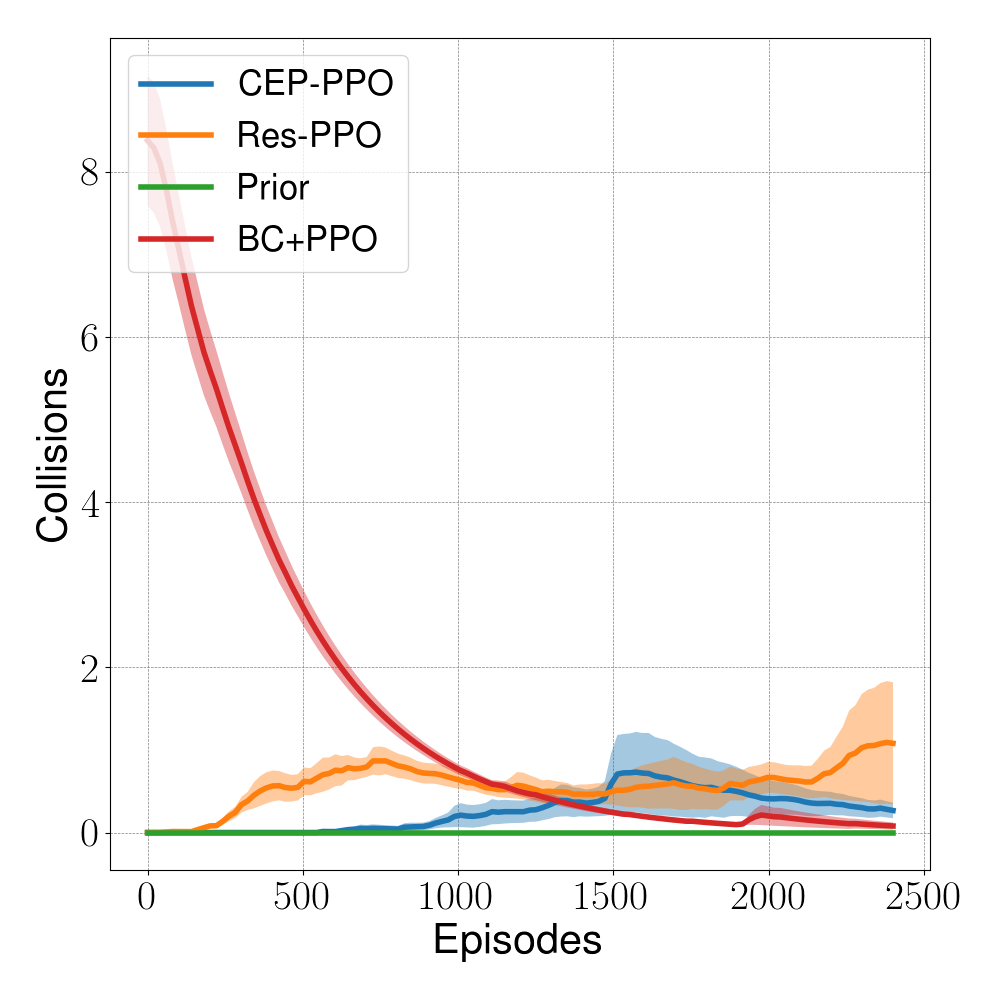}
	\end{minipage}
	\caption{\scriptsize{Training curves for Hitting a puck environment. \gls{cep}-PPO performs consistenly better than other prior + \gls{rl} methods. Collision avoidance prior is stronger in \gls{cep}-PPO and provides a safer training.}}
	\label{fig:rl_solution}
	\vspace{-0.5cm}
\end{figure}

\paragraph{\textbf{Results and analysis}}
We show the obtained results in Fig.~\ref{fig:rl_solution}. Our method, CEP-PPO performs on average better than Residual Learning and Behavioural Cloning + RL methods. The obtained results show that while \gls{cep} can consistently improve its performance, residual policy learning has a slower learning curve and behavioral cloning + RL decays in performance. We remark that the three methods start performing worse than the prior due to the stochasticity of the policies in contrast with the deterministic prior. The obtained results might be related to the collisions per episode. \gls{cep} based policy allows imposing hard constraints to avoid collisions. Both, behavioral cloning + RL and residual learning adds white noise around the optimal prior action and then, increase the chances of table collision. Fewer collisions mean that $r_T(\vx_{ee})$ is less activated during the training and then, the learning can focus on moving the puck close to the target. In the Behavioural Cloning case, we see that the table collisions push the robot to avoid the table, and then, it forgets how to reach the target. The Behavioural Cloning + RL policy ends with a policy that makes the robot neither collide with the table and neither to hit the puck. In the residual learning case, the policy is updated by taking both hitting the puck and avoid the table into consideration, while in \gls{cep}, the collisions against the table are less, and thus, the training is more focused on improving the hitting tactic.

\begin{figure}
	\centering
	\begin{minipage}{.24\textwidth}
		\centering
		\includegraphics[width=.99\textwidth]{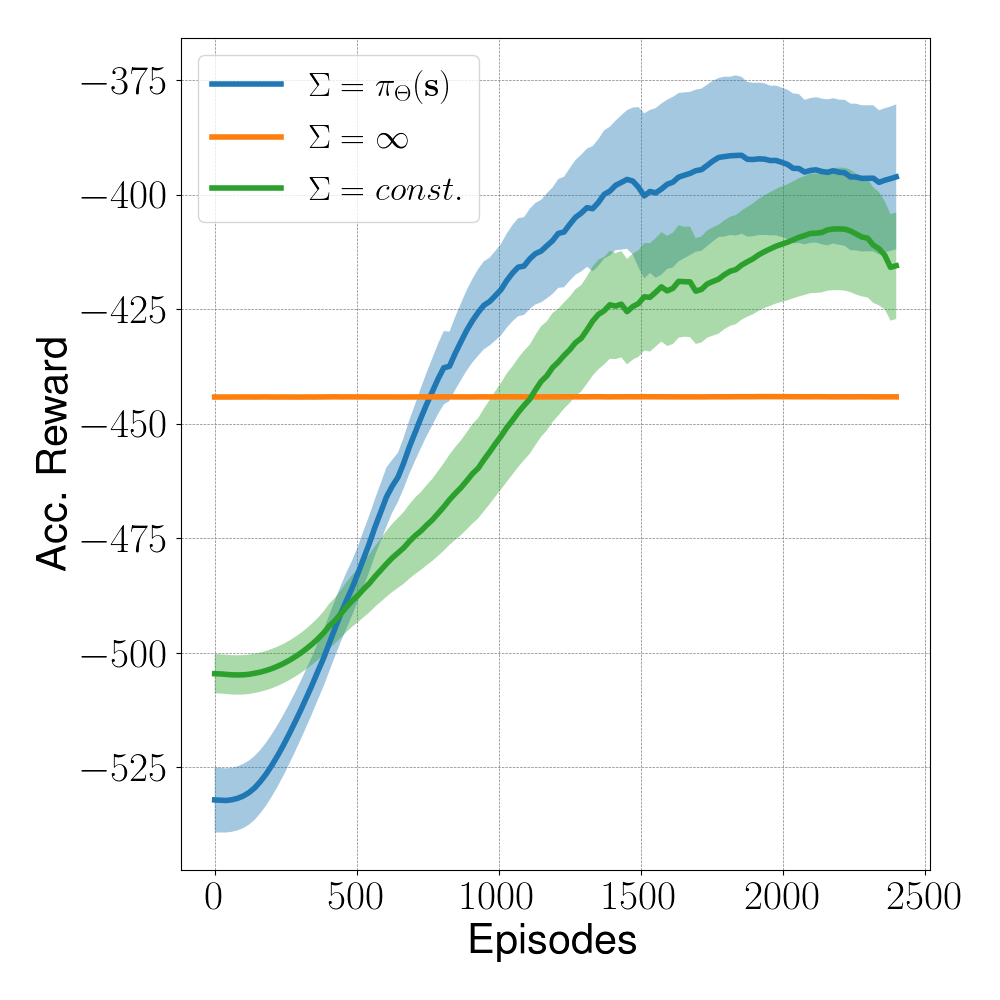}
	\end{minipage}
	\begin{minipage}{.24\textwidth}
		\centering
		\includegraphics[width=.99\textwidth]{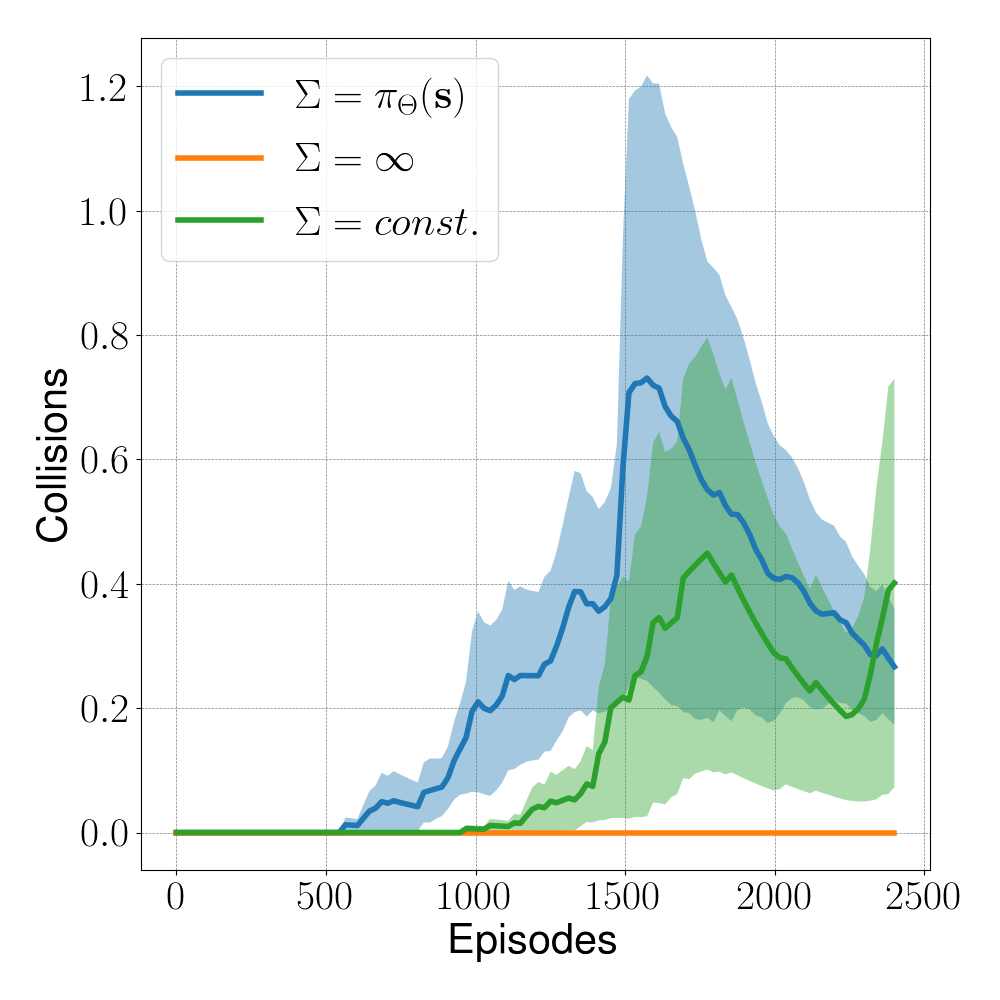}
	\end{minipage}
	\caption{\scriptsize{Training curves for Hitting a puck environment. Learning the variance of the policy improves the learning curve, but also increases the probability of collisions. }}
	\label{fig:rl_solution2}
\end{figure}

\paragraph*{\textbf{\gls{cep} ablation study}}
When applying CEP-PPO, the $\vSigma_{H}$ parameter from \eqref{eq:hrl_policy} controls how big is the influence of the learnable policy in the overall policy. The smaller $\vSigma_{H}$ is; the smaller the entropy for the learnable part and the higher the influence in the overall policy. In order to study the influence of the $\vSigma_{H}$ parameter, we compare CEP-PPO with fixed $\vSigma_{H}$ and CEP-PPO with learnable $\vSigma_{H}$. We present the obtained results in Fig.~\ref{fig:rl_solution2}
As expected having the freedom to set the $\vSigma_{H}$ parameter allows the robot to learn a better behavior to hit the puck. If the $\vSigma_{H}$ parameter is fix and too big, then the learned policy has a low impact on the overall policy and it can not improve a lot. With a very small $\vSigma_{H}$, the learned policy will have a very high impact in the overall policy and then, it might not get benefited by the priors. When learning $\vSigma_{H}$ parameter in the high-level policy, the robot decides when to trust more in the priors and when less. This allows the robot to get help from priors only if required, but also to decouple from them to learn a better policy. However, shown in the collisions plot from Fig.~\ref{fig:rl_solution2}, reducing the effect of the priors might increase the probability to go into dangerous regions and collide more often against the table. Nevertheless, learning the $\vSigma_{H}$ parameter seems to provide better results than keeping it fixed. 

\section{Conclusions}
We have presented Composable Energy Policies, a novel approach for modular robot control. In contrast with the previous method for reactive motion generation, which computes the optimal action for each component independently; our method solves an optimization problem over the sum of components. We have shown that \gls{apf} policies can be rewritten as a \gls{cep} and provide reasoning of why their method might fail. Through an evaluation phase, we have also shown that our method can solve a set of reactive motion generation problems more efficiently than artificial potential field methods.

Our method is flexible and allows us to represent each component in an arbitrary set of state-action spaces. Additionally, \gls{cep} allows integrating policies from multiple sources and samples from the product of them. In our work, we show how to integrate a \gls{rl} agent with a set of priors. The resulting algorithm combines a high-level \gls{rl} policy with a low-level optimization problem. This low-level optimization allows to guarantee that the policy is safe (if an obstacle avoidance prior is added) or to explore in informative regions (if guiding priors are added). We have compared our method with two methods that also combine prior information with an \gls{rl} agent. Given the low-level optimization of our method, we can guarantee that some priors such as obstacle avoidance were satisfied and the learning is more stable.

Additionally, we have introduced in the Appendix, the relations between optimal control and reactive motion generation. Setting their connections allows us to compute how the optimal policy for a multi-objective optimal control problem diverges from the product of the optimal policies for each objective. These insights allow us to better understand the policy composition and improve the design of our components.

\newpage

\bibliographystyle{plainnat}
\bibliography{main}

\clearpage

\begin{appendices}

\section{A Control as Inference view for Reactive motion generation}
\label{sec:optimality}
In the following section, we want to highlight the connections between reactive motion generation and control as inference to evaluate the optimality guarantees of composing energies in a multi-objective optimal control problem.

\begin{figure}[ht]
	\caption{Graphical model for the Optimal Control problem. $\vs_t$ denoted the state, $\va_t$ denotes the action and $o_t$ is an additional variable representing the optimality of the state and action for a given reward.}
	\begin{center}
			\resizebox{0.45\textwidth}{!}{\input{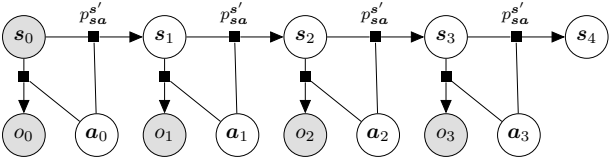}}
	\end{center}
	\label{fig:factor_graph_0}
\end{figure}

We frame optimal control as Bayesian inference problem~\cite{rawlik2012stochastic,levine2018reinforcement} over the sequence of actions $\va_{0:T}$. The optimal control problem is visualized as a graphical model in Fig.~\ref{fig:factor_graph_0}. The problem is formulated introducing an auxiliary variable $o_{0:T}$ that represents the optimality of $\vs_t$ and $\va_t$ under a certain reward function, $p(o_{t}|\vs_t, \va_t) \propto \exp(r(\vs_t,\va_t))$. Given a certain prior distribution $q(\va|\vs_0)$ and given $\vs_0$ is known, the inference problem is
\begin{align}
	\label{eq:control_as_inference}
	p(A_{0}^{T} | \vs_0, O_{0}^{T} ) = \frac{p(O_{0}^{T}|A_{0}^{T},\vs_0) q(A_{0}^{T}|\vs_0)}{p(O_{0}^{T}|\vs_0)}
\end{align}
with
\begin{align}
	p(O_{0}^{T}|A_{0}^{T},\vs_0) = \int_{\vs_{1:T}} p(O_{0}^{T}|S_{0}^{T},A_{0}^{T})p(S_{1}^{T}|A_{0}^{T},\vs_0)d S_{1}^{T}
\end{align}
where $A_{0}^{T}:\{\va_0,\dots, \va_T\}$, $O_{0}^{T}:\{o_0,\dots, o_T\}$ and  $S_{0}^{T}:\{\vs_0,\dots, \vs_T\}$.
The are two main directions to solve the posterior in \eqref{eq:control_as_inference}. First, methods that frame the problem as an \gls{hmm} and solve it in an Expectation-Maximization approach. Second, methods that compute the posterior in the trajectory level $A_{0}^{T}$. The first, are computationally demanding as they require several forward and backward message passing to compute the posterior. The second, needs to solve the problem in the trajectory level and thus the dimension of the variables grows linearly with the trajectory length, $T$.

Reactive motion generation instead, solves a one-step-ahead problem. The solutions in reactive motion generation are local rather than global and thus, they are computationally more efficient. 

\begin{figure}[ht]
	\caption{Graphical model for one-step ahead optimal control problem. In this approach, the actions $A_{1}^{T}$ are dependant on $S_{1}^{T}$ given a policy $\pi$.}
	\begin{center}
			\resizebox{0.45\textwidth}{!}{\input{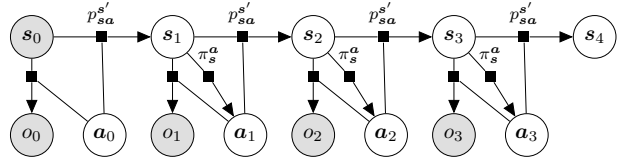}}
	\end{center}
	\label{fig:one_step_optimal}
\end{figure}

We frame reactive motion generation as a one-step control as inference problem
\begin{align}
	p( \va_{0} | \vs_0, O_{0}^{T} ) = \frac{p(O_{0}^{T}|\va_{0},\vs_0)q(\va_{0}|\vs_0)}{p(O_{0}^{T}|\vs_0)}
\end{align}
with
\begin{align}
	&p(O_{0}^{T}|\va_0,\vs_0) = \nonumber \\
	&\int_{S_{1:T}}\int_{A_{1:T}} p(O_{0}^{T}|S_{0}^{T},A_{0}^{T})p(S_{1}^{T}|\va_0,\vs_0) \pi(A_{1}^{T}|S_{1}^{T})d S_{1}^{T} d A_{1}^{T}.
	\label{eq:adapted_op}
\end{align}
In contrast with the classical stochastic optimal control problem that finds the posterior for the whole trajectory $A_{0}^{T}$, in one-step-ahead control as inference problem, we aim to find the posterior for only the instant next control action, $\va_0$. Computing the posterior only for $\va_0$ requires the likelihood to be defined as the marginal of not only the state trajectory $S_{1}^{T}$, but also the action trajectory $A_{1}^{T}$. The graphical model for one-step-ahead control as inference is presented in Fig.~\ref{fig:one_step_optimal}.
In one-step ahead control as inference, we introduce an additional distribution $\pi$ that provides us the probability of $A_{1}^{T}$ given $S_1^T$. This additional policy $\pi$ is interpreted as the policy the agent will apply in the future. In this context, given we know the policy the agent will apply in the future, we are looking for the instant action the maximized the cumulative reward in the long horizon trajectory. \eqref{eq:adapted_op} can be rewritten as the expectation over $p(O_{0}^{T}|S_{0}^{T},A_{0}^{T})$
\begin{align}
	\E_{S_{0}^{T},A_{0}^{T}\sim p^{\pi}(S_{0}^{T},A_{0}^{T}|\vs_0,\va_0)}\left[  p(O_{0}^{T}|S_{0}^{T},A_{0}^{T}) \right] \propto \nonumber \\
	\E_{S_{0}^{T},A_{0}^{T}\sim p^{\pi}(S_{0}^{T},A_{0}^{T}|\vs_0,\va_0)} \left[ \exp (\sum_{t=0}^{T} r(\vs_t, \va_t))\right]= \nonumber \\
	\exp(Q_{r}^{\pi}(\vs_0, \va_0))
\end{align}
From what follows, the likelihood for our inference problem is proportional to the $\exp(Q)$ defined over a certain policy $\pi$. Given a $\pi$, the posterior for our inference problem is given by
\begin{align}
	p(\va_0|\vs_0, O_0^T) \propto 	\exp(Q_{r}^{\pi}(\vs_0, \va_0))q(\va_{0}|\vs_0).
\end{align}

The provided $Q$ function depends on $\pi$ and then, the quality of our reactive motion generator to solve a long horizon optimal control problem directly depends on the quality of $\pi$. In the optimal case, for $Q^*$, the reactive motion generator can find the optimal trajectory, even if the optimization is done locally.

\subsection{Optimality Guarantees}

In \gls{cep}, we propose to model the $Q$ function as the sum of a set of optimal $Q_k^*$ functions.
Instead, we are aware that $Q_\Sigma = \frac{1}{K}\sum_{k=0}^{K} Q_k^*$ is not the optimal function $Q^*$ for the sum of the rewards $r = \frac{1}{K} \sum_{k=1}^{K}r_k$. The closer $Q_\Sigma$ is from the optimal $Q^*$, the closer the product of experts policy would be from the optimal policy. In this section, we study, given a certain reward $r = \frac{1}{2} (r_1 + r_2)$, how much the sum of the individual components $Q_\Sigma$ diverge from the optimal $Q^*$. In \cite{levine2018reinforcement} is shown, that the optimal Q function for the control as inference problem can be computed by recursively solving the soft-value iteration~\cite{ziebart2010modeling}
\begin{align}
	Q(\vs,\va) &= r(\vs,\va) + \E_{p(\vs'|\vs,\vs)}\left[V(\vs')\right] \nonumber\\
	V(\vs) &= \log \int_\gA \exp(Q(\va,\vs)).
\end{align}
We assume a finite horizon control as inference problem and evaluate how much $Q_\Sigma$ diverges from $Q*$ when increasing the control horizon

For $t=T$,
\begin{align}
	Q^{*T} = \frac{1}{2} (r_1 + r_2)
\end{align}
and
\begin{align}
	Q_{1}^{*T} = r_1 \, , \,
	Q_{2}^{*T} = r_2.
\end{align}
Then, 
\begin{align}
	\label{eq:T_step_sol}
	Q^{*T} = \frac{1}{2}(Q_{1}^{*T} + Q_{2}^{*T})
\end{align}
and the divergence 
\begin{align}
	\Delta Q^{T} = Q^{*T} - Q_{\Sigma}^{T} = 0.
\end{align}
From \eqref{eq:T_step_sol}, for $T=1$ horizon optimal control problems, the sum of the optimal components $Q_\Sigma$ is equal to the optimal $Q$. For longer horizon control as inference problems, the optimal $Q$ is computed by recursively solving the soft-bellman update backward in time. For computing $t=T-1$,
\begin{align}
	Q^{*T-1}(\vs,\va) &= r(\vs,\va) + \E_{p(\vs'|\vs,\vs)}\left[V^{*T}(\vs')\right] \nonumber \\
	V^{*T}(\vs) &= \log \int_\gA \exp(Q^{*T}(\va,\vs)).
\end{align}
we do a soft bellman update. The difference betwen $Q^{*T-1}$ and $Q_{\Sigma}^{T-1}$
\begin{align}
	\label{eq:error_01}
	\Delta Q^{T-1} = Q^{*T-1} - Q_{\Sigma}^{T-1} = \E_{p(\vs'|\vs,\va)}\left[ V^{*T} - V_{\Sigma}^{T}\right] \nonumber \\
\end{align}
with
\begin{align}
	\label{eq:error_02}
	V^{*T} - V_{\Sigma}^{T} = \log \frac{\int_\gA \exp(Q^{*T})}{\int_\gA (\exp(Q_1^{*T})\int_\gA\exp(Q_2^{*T}))^{\frac{1}{2}}} \nonumber \\
	= \log \frac{\int_\gA \exp(\frac{1}{2}Q_1^{*T})\exp(\frac{1}{2}Q_2^{*T})   }{(\int_\gA \exp(Q_1^{*T})\int_\gA\exp(Q_2^{*T}))^{\frac{1}{2}}}.
\end{align}
Then,
\begin{align}
	Q^{*T-1} &= Q_{\Sigma}^{T-1} + \Delta Q^{T-1} \nonumber\\
	&= Q_{\Sigma}^{T-1} + \E_{p(\vs'|\vs,\va)}\left[\log \frac{\int_\gA \exp(\frac{1}{2}Q_1^{*T})\exp(\frac{1}{2}Q_2^{*T})   }{(\int_\gA \exp(Q_1^{*T})\int_\gA\exp(Q_2^{*T}))^{\frac{1}{2}}}  \right]
\end{align}

From \eqref{eq:error_01} and \eqref{eq:error_02}, we can obtain the recurrence relation for the divergence error
\begin{align}
	\label{eq:recurrence_for_delta}
	\Delta Q^{t-1} = \E_{p(\vs'|\vs,\va)}\left[\log \frac{\int_\gA \exp(\frac{1}{2}Q_1^{*t})\exp(\frac{1}{2}Q_2^{*t})\exp(\Delta Q^{t}) }{(\int_\gA \exp(Q_1^{*t})\int_\gA\exp(Q_2^{*t}))^{\frac{1}{2}}}  \right].
\end{align}
From \eqref{eq:error_01} and \eqref{eq:error_02}, we can see that if $Q_1^* = Q_2^*$, then $\Delta Q =0$ and the more they differ, the bigger the divergence error to the optima $Q^*$.

We can also extrapolate interesting results from \eqref{eq:recurrence_for_delta}. For any $Q_1^{*}$ and $Q_2^{*}$,
\begin{align}
	(\int_\gA \exp(Q_1^{*t})\int_\gA\exp(Q_2^{*t}))^{\frac{1}{2}} > \int_\gA \exp(\frac{1}{2}Q_1^{*t})\exp(\frac{1}{2}Q_2^{*t})
\end{align}
and then,
\begin{align}
	\label{eq:delta_error}
	\Delta Q < 0 , \forall \, Q_1^*,Q_2^*
\end{align}
And then, from \eqref{eq:recurrence_for_delta}, $\Delta Q$ monotonically  decreases. The longer the horizon T, the bigger the divergence error $\Delta Q$. 

In practice, these insights are beneficial for modeling our $Q$ functions. We can expect the performance of our reactive motion generation to decay in those environments in which long-horizon planning is required as the $\Delta Q$ will increase over long horizons. On the other hand, the more overlapping regions the $Q$ components have, the smaller the divergence error would be and the closer we would be to the optima.

Similar theoretical studies have been already developed~\cite{haarnoja2018composable, van2019composing, todorov2009compositionality}. In Optimal Control, composable optimality guarantees were proven for linear dynamics, by the \gls{lmdp} approach. In \cite{todorov2009compositionality, da2009linear}, a weighted policy sum was proven to be optimal for the sum of the rewards, as long as the rewards differ only in the terminal reward.

In \cite{haarnoja2018composable, tasse2020boolean}, the optimality of the composition is studied in a maximum entropy reinforcement learning problem
\begin{align}
	J(\pi) =  \E_{\pi,p(\vs_0)}\left[\sum_{t=0}^T r(\vs_t,\va_t) + \alpha \gH(\pi)| \vs_{t+1} \sim p(\cdot|\vs_t, \va_t)\right]
\end{align}
with $\gH(\pi)$ the entropy of the policy. In \cite{haarnoja2018composable} is proven, that optimal soft-Q function for $r = \frac{1}{2}r_1+r_2$ is bounded
\begin{align}
	Q_{\Sigma}(\vs,\va) \geq Q^{*}(\vs,\va) \geq Q_{\Sigma}(\vs,\va) - \gC^*(\vs,\va)
\end{align}
with $Q_{\Sigma} = \frac{1}{2} Q_1^* + Q_2^*$ and $\gC^*$ is the fixed point of
\begin{align}
	\gC(\vs,\va) \leftarrow \gamma \E_{p(\vs'|\vs,\va)}\left[\gD_{\frac{1}{2}}(\pi_1^*(\va|\vs)||\pi_2^*(\va|\vs)) + \max_{\va'} \gC(\vs',\va') \right]
\end{align}
and $\gD_{\frac{1}{2}}$ is the Renyi divergence of order $\frac{1}{2}$.

\section{Composable Energy Policies as Inference}
\label{sec:inference_cep}
\begin{figure}[ht]
	\caption{Graphical model for Composable Energy Policies. $o_k$ is an auxiliary variable that represents the optimality of $\vs_0$ and $\va_0$ for a particular energy.}
	\begin{center}
		\input{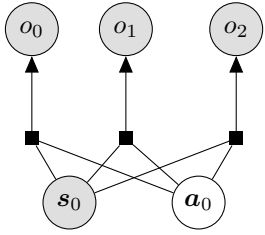}
	\end{center}
	\label{fig:cep_inference}
\end{figure}

The product of experts is the natural expression for the multi-objective inference problem presented in Fig.~\ref{fig:cep_inference}. We introduce a set of auxiliary variables $o_0,\dots, o_K$ that represents the optimality of $\vs_0$ and $\va_0$ given a certain energy function $p(o_k=1|\vs_0,\va_0) \propto \exp(E_k(\vs,\va))$. 
Then, the likelihood for the graphical model in Fig.~\ref{eq:cep_generic} can be computed as the product of the terms
\begin{align}
    p(\vs_0,\va_0, o_{0:2}) = q(\va_0)p(\vs_0)\prod_{k=0}^{2}p(o_k|\vs_0,\va_0).
\end{align}
We are interested on computing the maximum a posteriori for $\va_0$ given $\vs_0$ and $o_k=1\,,\, \forall \,k=(0,1,2)$
\begin{align}
    \va_0^* &= \argmax_{\va_0} p(\va_0|\vs_0,o_{0:2} =1) \nonumber \\
    &\propto \exp\left(\sum_{k=0}^2 E_k(\va_0,\vs_0)\right)q(\va_0).
    \label{eq:maximum_a_posterior}
\end{align}
From \eqref{eq:maximum_a_posterior}, we can see that the maximum a posteriori will search for the optimal $\va_0$ given an exponentiated sum of energies and the prior distribution $q(\va_0)$. If we assume a uniform distribution for the prior, we recover \eqref{eq:product_of_energies}. Thus, the product of experts represents the distribution that satisfies best a set of components $E_0,\dots, E_k$. We highlight that the \gls{cep} graphical model in Fig.~\ref{fig:cep_inference} and in Fig.~\ref{fig:energy_tree} are equivalent. The model in Fig.~\ref{fig:cep_inference},  represent the \gls{cep} as the distribution maximizing a set of rewards $E_0,\dots\ E_k$, while the model in Fig.~\ref{fig:energy_tree} represent the \gls{cep} as the distribution maximizing the combination of a set of policies. Depending on the problem, we might be interested to frame the problem with one or the other approach.

\section{Composable Energy Policies and Artificial Potential Fields Method}
\label{sec:prove_2}

In the following section, we will show that the Artificial Potential Fields Method can be written as a particular case of Composable Energy Policies.

Artificial Potential Field Method assumes a weighted sum of the accelerations generated by a set of dynamic components
\begin{align}
	\label{eq:apf_appendix}
	\ddot{\vq} = \sum_{k=0}^{K} \vLambda_k \vg_k(\vq, \dot{\vq}).
\end{align}
Let's consider a particular case of \gls{cep} were all the components are represented with a normal distribution centered in $\vg_k(\vq, \dot{\vq})$,
\begin{align}
	\pi_k(\ddot{\vq}|\vq, \dot{\vq}) = \gN(\vg_k(\vq, \dot{\vq}), \Sigma_k).
\end{align}
Then, given $K$ components
\begin{align}
	\label{eq:gaussians}
	\pi(\ddot{\vq}|\vq, \dot{\vq}) &= \prod_{k=0}^{K}\gN(\vg_k(\vq, \dot{\vq}), \Sigma_k) = \gN(\vg_\Sigma(\vq, \dot{\vq}) , \Sigma_\Sigma)
\end{align}
with
\begin{align}
		\label{eq:mean}
		\vg_\Sigma = \left(\sum_{k=0}^{K} \Sigma_{k}^{-1} \right)^{-1} \left(\sum_{k=0}^{K} \Sigma_{k}^{-1}\vg_{k}\right)
\end{align}
and
\begin{align}
		\Sigma_\Sigma = \left( \sum_{k=0}^{K} \Sigma_{k}^{-1} \right)^{-1}.
\end{align}
In \gls{cep}, the acceleration is obtained by an optimization of
\begin{align}
	\ddot{\vq} = \argmax_{\ddot{\vq}} \pi(\ddot{\vq}|\vq, \dot{\vq}).
\end{align}
The maximum of \ref{eq:gaussians} is known; given the policy is a gaussian, the maximun is the mean in \ref{eq:mean}
\begin{align}
	\label{eq:max_cep_gauss}
	\ddot{\vq} = \left(\sum_{k=0}^{K} \Sigma_{k}^{-1} \right)^{-1} \left(\sum_{k=0}^{K} \Sigma_{k}^{-1}\vg_{k}(\vq,\dot{\vq})\right).
\end{align}
We can rewrite \eqref{eq:max_cep_gauss} to \eqref{eq:apf_appendix} 
\begin{align}
	\vLambda_k = \left(\sum_{k=0}^{K} \Sigma_{k}^{-1} \right)^{-1}. \Sigma_{k}^{-1} 
\end{align}
We show that \gls{apf} policy can be rewritten as a \gls{cep}, modelling the dynamics by a product of normal distributions. Modelling all the components by normal distributions might be a bad choice for a proper modules integration. normal distributions assume that (1) there is a unique optimal action (the mean) to solve the task and (2) the actions quality is related to the mahalanobis distance to the optima. While tasks like reaching a target might satisfy (1) and (2); tasks like obstacle avoidance might require richer representations to properly solve the task.

\section{Experiments}
\label{sec:experiments}

\subsection{Reactive Motion Generation}
The modular components (Go-to, Obstacle Avoidance and joint limits avoidance) for both baselines \gls{apf} and \gls{rmp} were modeled based on \cite{khatib1985potential} and \cite{cheng2018rmpflow} respectively. \gls{cep} were modeled following the energies presented in section~\ref{sec:robot_motion}. We introduce a table with the hyperparameters for all \gls{cep} energies
\begin{table}[ht]
	\centering
	\resizebox{0.45\textwidth}{!}{
		\begin{tabular}{c c c c} 
			\hline
			\textbf{Energy Modules} & \multicolumn{3}{c}{\textbf{Parameters}}\\
			\hline
			Reach Target & $\mK_p = 20.$ & $\mK_v=30.$ & $\alpha=10.$ \\
			Obstacle Avoidance &  $\gamma = 0.2$ & $\alpha = 4.$ & $\beta = 0.1$ \\
			Joint Limits avoidance & $\gamma = 0.3$ & $\alpha = 4.$ & $\beta = 0.1$  \\
			\hline
		\end{tabular}}
		\vspace{0.1cm}
	\caption{Component parameters for Composable Energy Policies in reactive motion generation}
	\label{tab:cep_energies}
\end{table}

\subsection{Prior based policies for RL}

We introduce a Table with the Hyperparameters for the PPO in Mushroom-RL~\cite{deramo2020mushroomrl}. We used the same hyperparameters for the three methods (Behavioural Cloning + \gls{rl}, Residual \gls{rl} and \gls{cep} for \gls{rl}).
\begin{table}[ht]
	\centering
	\resizebox{0.3\textwidth}{!}{
		\begin{tabular}{c c} 
			\hline
			\textbf{Hyperparameters} & \\
			\hline
			horizon & 1400\\
			Policy Net &  18-128-128-3\\
			policy std0 & 1. \\
			policy batch & 256 \\
			n\_epochs\_policy & 2\\
			policy learn rate & 1e-4 \\
			Critic Net &  18-256-256-1\\
			critic batch & 256\\
			critic learn rate & 3e-4\\
			n\_steps\_per\_fit & horizon x 30 \\
			Discount factor & .997 \\
			\hline
		\end{tabular}}
		\vspace{0.1cm}
	\caption{Component parameters for Composable Energy Policies in reactive motion generation}
	\label{tab:priors}
\end{table}

\end{appendices}

\end{document}